\documentclass{article}

\usepackage{times}
\usepackage{graphicx} % more modern
\usepackage{subfigure} 

\usepackage{natbib}

\usepackage{algorithm}
\usepackage{algorithmic}

\usepackage{amsmath,amsfonts,amssymb}
\newcommand\norm[1]{\left\lVert#1\right\rVert}
\usepackage{booktabs}
\usepackage{footnote}
\makeatletter
\def\@xfootnote[#1]{%
  \protected@xdef\@thefnmark{#1}%
  \@footnotemark\@footnotetext}
\makeatother

\usepackage{hyperref}

\usepackage[accepted]{icml2016}

\icmltitlerunning{Neural Graph Machines: Learning Neural Networks Using Graphs}

\usepackage{cleveref}
\begin{document} 

\twocolumn[
\icmltitle{Neural Graph Machines: Learning Neural Networks Using Graphs}
% proposed new title Neural Networks meet Graphs: Revisiting Graph Augmentation

% proposed new title: Neural Gragh Machines: Revisiting neural network training using graph augmentation. 



% you can specify symbols, otherwise they are numbered in order

\icmlauthor{Thang D. Bui$^*$\footnotemark}{tdb40@cam.ac.uk}
\icmlauthor{Sujith Ravi$^\dagger$}{sravi@google.com}
\icmlauthor{Vivek Ramavajjala$^\dagger$}{vramavaj@google.com}
\vskip 0.1in
\icmladdress{$^*$ University of Cambridge, United Kingdom}
\vskip -0.15in
\icmladdress{$^\dagger$ Google Research, Mountain View, CA, USA}

\vskip 0.3in
]

\footnotetext{Work done during an internship at Google.}

\begin{abstract}
Label propagation is a powerful and flexible semi-supervised learning technique on graphs. Neural networks, on the other hand, have proven track records in many supervised learning tasks. In this work, we propose a training framework with a graph-regularised objective, namely {\it Neural Graph Machines}, that can combine the power of neural networks and label propagation. This work generalises previous literature on graph-augmented training of neural networks, enabling it to be applied to multiple neural architectures (Feed-forward NNs, CNNs and LSTM RNNs) and a wide range of graphs. The new objective allows the neural networks to harness both labeled and unlabeled data by: (a)~allowing the network to train using labeled data as in the supervised setting, (b)~biasing the network to learn similar hidden representations for neighboring nodes on a graph, in the same vein as label propagation. Such architectures with the proposed objective can be trained efficiently using stochastic gradient descent and scaled to large graphs, with a runtime that is linear in the number of edges. The proposed joint training approach convincingly outperforms many existing methods on a wide range of tasks (multi-label classification on social graphs, news categorization, document classification and semantic intent classification), with multiple forms of graph inputs (including graphs with and without node-level features) and using different types of neural networks.
\end{abstract}
\section{Introduction}
\label{sec:intro}
Semi-supervised learning is a powerful machine learning paradigm that can improve the prediction performance compared to techniques that use only labeled data, by leveraging a large amount of unlabeled data. The need of semi-supervised learning arises in many problems in computer vision, natural language processing or social networks, in which getting labeled datapoints is expensive or unlabeled data is abundant and readily available. 
 
There exist a plethora of semi-supervised learning methods. The simplest one uses bootstrapping techniques to generate pseudo-labels for unlabeled data generated from a system trained on labeled data. However, this suffers from label error feedbacks \citep{lee2013pseudo}. In a similar vein, autoencoder based methods often need to rely on a two-stage approach: train an autoencoder using unlabeled data to generate an embedding mapping, and use the learnt embeddings for prediction. In practice, this procedure is often costly and inaccurate. Another example is transductive SVMs \citep{joachims1999transductive}, which is too computationally expensive to be used for large datasets. Methods that are based on generative models and amortized variational inference \citep{kingma2014semi} can work well for images and videos, but it is not immediately clear on how to extend such techniques to handle sparse and multi-modal inputs or graphs over the inputs. 

In contrast to the methods above, graph-based techniques such as label propagation \citep{zhu2002learning, bengio2006label} often provide a versatile, scalable, and yet effective solution to a wide range of problems. These methods construct a smooth graph over the unlabeled and labeled data. Graphs are also often a natural way to describe the relationships between nodes, such as similarities between embeddings, phrases or images, or connections between entities on the web or relations in a social network. Edges in the graph connect semantically similar nodes or datapoints, and if present, edge weights reflect how strong such similarities are. By providing a set of labeled nodes, such techniques iteratively refine the node labels by aggregating information from neighbours and propagate these labels to the nodes' neighbours. In practice, these methods often converge quickly and can be scaled to large datasets with a large label space \citep{ravi2016large}. We build upon the principle behind label propagation for our method.

Another key motivation of our work is the recent advances in neural networks and their performance on a wide variety of supervised learning tasks such as image and speech recognition or sequence-to-sequence learning \citep{krizhevsky2012imagenet, hinton2012deep, sutskever2014sequence}. Such results are however conditioned on training very large networks on large datasets, which may need millions of labeled training input-output pairs. This begs the question: can we harness previous state-of-the-art semi-supervised learning techniques, to jointly train neural networks using limited labeled data and unlabeled data to improve its performance?
\vspace{-0.1in}
\paragraph{Contributions:} We propose a discriminative training objective for neural networks with graph augmentation, that can be trained with stochastic gradient descent and efficiently scaled to large graphs. The new objective has a regularization term for generic neural network architectures that enforces similarity between nodes in the graphs, which is inspired by the objective function of label propagation. In particular, we show that:
\begin{itemize}
	\item Graph-augmented neural network training can work for a wide range of neural networks, such as feed-forward, convolutional and recurrent networks. Additionally, this technique can be used in both inductive and transductive settings. It also helps learning in low-sample regime (small number of labeled nodes), which cannot be handled by vanilla neural network training.
	\item The framework can handle multiple forms of graphs, either naturally given or constructed based on embeddings and knowledge bases.
	\item Using graphs and neighbourhood information alone as direct inputs to neural networks in this joint training framework permits fast and simple inference, yet provides competitive performance with current state-of-the-art approaches which employ a two-step method of first training a node embedding representation from the graph and then using it as feature input to train a classifer separately (see \cref{sec:exp_graph}).
	\item As a by-product, our proposed framework provides a simple technique to finding smaller and faster neural networks that offer competitive performance with larger and slower non graph-augmented alternatives (see \cref{sec:exp_cnn}).
\end{itemize}

We experimentally show that the proposed training framework outperforms state-of-the-art or perform favourably on a variety of prediction tasks and datasets, involving text features and/or graph inputs and on many different neural network architectures (see \cref{sec:experiments}).

The paper is organized as follows: we first review some background and literature, and relate them to our approach in \cref{sec:background}; we then detail the training objective and its properties in \cref{sec:ngm}; and finally we validate our approach on a range of experiments in \cref{sec:experiments}.
\section{Background and related works}
\label{sec:background}
In this section, we will lay out the groundwork for our proposed training objective in section \ref{sec:ngm}.

\subsection{Neural network learning}

Neural networks are a class of non-linear mapping from inputs to outputs and comprised of multiple layers that can potentially learn useful representations for predicting the outputs. We will view various models such as feed-forward neural networks, recurrent neural networks and convolutional networks under the same umbrella. Given a set of $N$ training input-output pairs $\{x_n, y_n\}_{n=1}^{N}$, such neural networks are often trained by performing maximum likelihood learning, that is, tuning their parameters so that the networks' outputs are close to the ground truth under some criterion,
\begin{align}
\mathcal{C}_{\mathrm{NN}}(\theta) = \sum_{n} c(g_\theta(x_n), y_n), \label{eqn:cost_nn}
\end{align}
where $g_\theta(\cdot)$ denotes the overall mapping, parameterized by $\theta$, and $c(\cdot)$ denotes a loss function such as $l$-2 for regression or cross entropy for classification. The cost function $c$ and the mapping $g$ are typically differentiable w.r.t~$\theta$, which facilitates optimisation via gradient descent. Importantly, this can be scaled to a large number of training instances by employing stochastic training using minibatches of data. However, it is not clear how unlabeled data, if available, can be treated using this objective, or if extra information about the training set, such as relational structures can be used.

\subsection{Graph-based semi-supervised learning}
In this section, we provide a concise introduction to graph-based semi-supervised learning using {\it label propagation} and its training objective. Suppose we are given a graph $G=(V, E, W)$ where $V$ is the set of nodes, $E$ the set of edges and $W$ the edge weight matrix. Let $V_l, V_u$ be the labeled and unlabeled nodes in the graph. The goal is to predict a soft assignment of labels for each node in the graph, $\hat{Y}$, given the training label distribution for the seed nodes, $Y$. Mathematically, label propagation performs minimization of the following convex objective function, for $L$ labels, 
\begin{align}
\mathcal{C}_{\mathrm{LP}}(\hat{Y}) &= \mu_1 \sum_{v \in V_l} \norm{\hat{Y}_v - Y_v}_2^2 \nonumber \\ 
&\quad + \mu_2 \sum_{v \in V, u \in \mathcal{N}(v)} w_{u,v} \norm{\hat{Y}_v - \hat{Y}_u}_2^2 \nonumber \\ 
&\quad + \mu_3 \sum_{v \in V} \norm{\hat{Y}_v - U}_2^2, \label{eqn:cost_lp}
\end{align}
subject to $\sum_{l=1}^{L}\hat{Y}_{vl} = 1$, where $\mathcal{N}(v)$ is the neighbour node set of the node $v$, and $U$ is the prior distribution over all labels, $w_{u,v}$ is the edge weight between nodes $u$ and $v$, and $\mu_1$, $\mu_2$, and $\mu_3$ are hyperparameters that balance the contribution of individual terms in the objective. The terms in the objective function above encourage that: (a)~the label distribution of seed nodes should be close to the ground truth, (b)~the label distribution of neighbouring nodes should be similar, and, (c)~if relevant, the label distribution should stay close to our prior belief. This objective function can be solved efficiently using iterative methods such as the Jacobi procedure. That is, in each step, each node aggregates the label distributions from its neighbours and adjusts its own distribution, which is then repeated until convergence. In practice, the iterative updates can be done in parallel or in a distributed fashion which then allows large graphs with a large number of nodes and labels to be trained efficiently. \citet{bengio2006label} and \citet{ravi2016large} provide good surveys on the topic for interested readers.

There are many variants of label propagation that could be viewed as optimising modified versions of \cref{eqn:cost_lp}. For example, manifold regularization \cite{belkin2006manifold} replaces the label distribution $\hat{Y}$ by a Reproducing Kernel Hilbert Space mapping from input features. Similarly, \citet{weston2012deep} also employs such mapping but uses a feed-forward neural network instead. Both methods can be classified as inductive learning algorithms; whereas the original label propagation algorithm is transductive \cite{yang2016revisiting}. 

These aforementioned methods are closest to our proposed approach; however, there are key differences.  Our work generalizes previously proposed frameworks for graph-augmented training of neural networks (e.g., \citet{weston2012deep}) and extends it to new settings, for example, when there is only graph input and no features are available. Unlike the previous works, we show that the graph augmented training method can work with multiple neural network architectures (Feed-forward NNs, CNNs, RNNs) and on multiple prediction tasks and datasets using {\it natural} as well as {\it constructed} graphs. The experiment results (see \cref{sec:experiments}) clearly validate the effectiveness of this method in all these different settings, in both inductive and transductive learning paradigms. Besides the methodology, our study also presents an important contribution towards assessing the effectiveness of graph combined neural networks as a generic training mechanism for different architectures and problems, which was not well studied in previous work. 

More recently, graph embedding techniques have been used to create node embedding that encode local structures of the graph and the provided node labels \cite{perozzi2014deepwalk, yang2016revisiting}. These techniques target learning better node representations to be used for other tasks such as node classification. In this work, we aim to directly learn better predictive models from the graph. We compare our method to these two-stage (embedding + classifier) techniques in several experiments in \cref{sec:experiments}.

% The embedding function can be learned separately, or jointly trained with the classifier mapping. 

Our work is also different and orthogonal to recent works on using neural networks {\it on} graphs, for example: \citet{defferrard2016convolutional} employs spectral graph convolution to create a neural-network like classifier. However, these approaches requires many approximations to arrive at a practical implementation. Here, we advocate a training objective that {\it uses} graphs to augment neural network learning, and works with many forms of graphs and with any type of neural network.

\section{Neural graph machines}
\label{sec:ngm}
\begin{figure*}[ht!]
\centering
\includegraphics[width=0.9\textwidth]{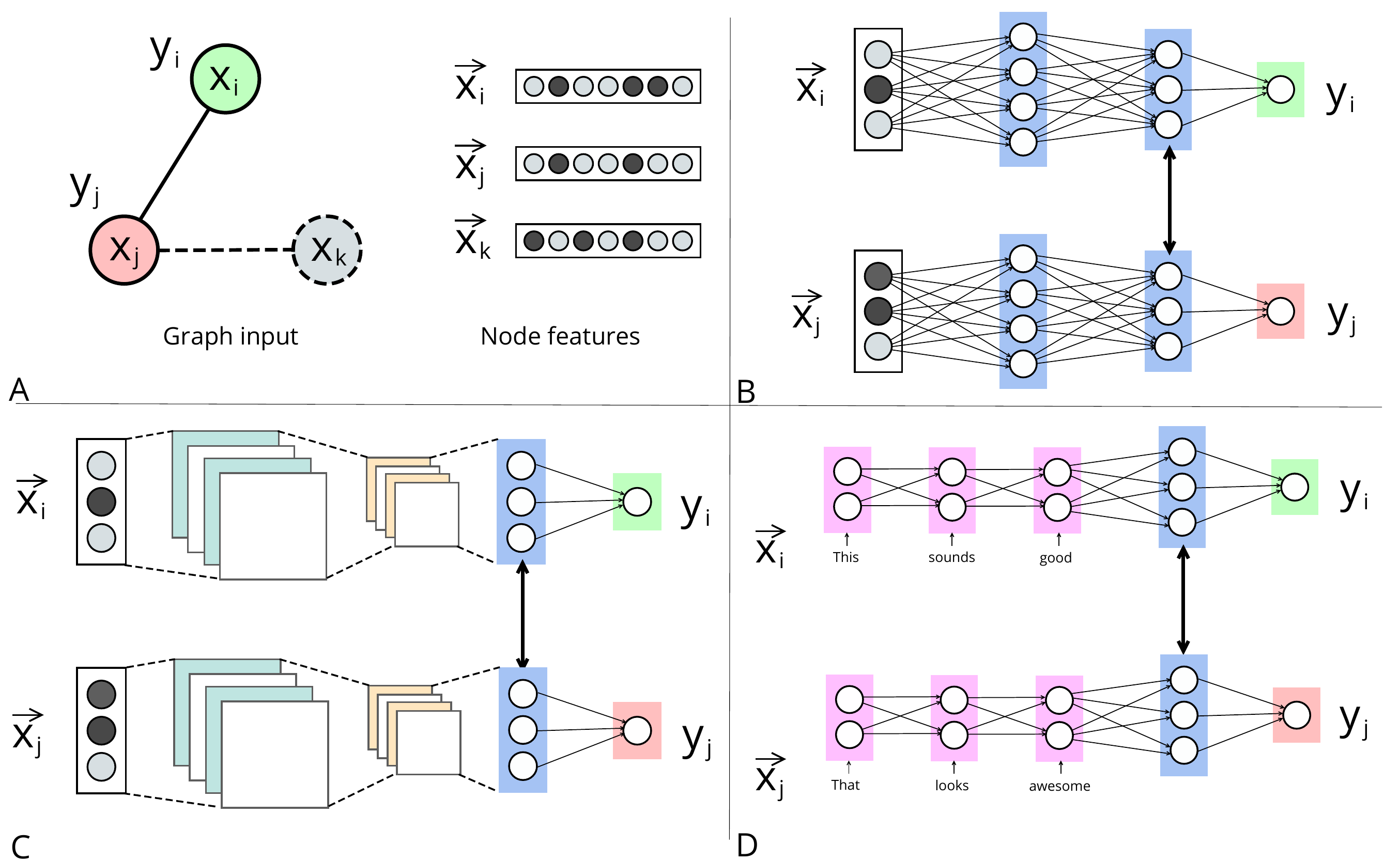}
\caption{A: An example of a graph and feature inputs. In this case, there are two labeled nodes ($x_i$, $x_j$) and one unlabeled node ($x_k$), and two edges. The feature vectors, one for each node, are used as neural network inputs. B, C and D: Illustration of Neural Graph Machine for feed-forward, convolution and recurrent networks respectively: the training flow ensures the neural net to make accurate node-level predictions and biases the hidden representations/embeddings of neighbouring nodes to be similar. In this example, we force $h_i$ and $h_j$ to be similar as there is an edge connecting $x_i$ and $x_j$ nodes.\label{fig:ngm}}
\end{figure*}
In this section, we devise a discriminative training objective for neural networks, that is inspired by the label propagation objective and uses both labeled and unlabeled data, and can be trained by stochastic gradient descent. 

First, we take a close look at the two objective functions discussed in section \ref{sec:background}. The label propagation objective equation \ref{eqn:cost_lp} ensures the predicted label distributions of neighbouring nodes to be similar, while those of labeled nodes to be close to the ground truth. For example: if a {\it cat} image and a {\it dog} image are strongly connected in a graph, and if the {\it cat} node is labeled as {\it animal}, the predicted probability of the {\it dog} node being {\it animal} is also high. In contrast, the neural network training objective equation \ref{eqn:cost_nn} only takes into account the labeled instances, and ensure correct predictions on the training set. As a consequence, a neural network trained on the {\it cat} image alone will not make an accurate prediction on the {\it dog} image.

Such shortcoming of neural network training can be rectified by biasing the network using prior knowledge about the relationship between instances in the dataset. In particular, for the domains we are interested in, training instances (either labeled or unlabeled) that are connected in a graph, for example, {\it dog} and {\it cat} in the above example, should have similar predictions. This can be done by encouraging neighboring data points to have a similar hidden representation learnt by a neural network, resulting in a modified objective function for training neural network architectures using both labeled and unlabeled datapoints. We call architectures trained using this objective {\it Neural Graph Machines}, and schematically illustrate the concept in figure \ref{fig:ngm}. The proposed objective function is a weighted sum of the neural network cost and the label propagation cost as follows,
\begin{align}
\mathcal{C}_{\mathrm{NGM}}(\theta) &= \sum_{n=1}^{V_l} c (g_\theta(x_n), y_n) \nonumber \\ 
& \quad + \alpha_1 \sum_{(u,v)\in \mathcal{E}_{LL}} w_{uv} d(h_\theta(x_u), h_\theta(x_v)) \nonumber \\ 
& \quad + \alpha_2 \sum_{(u,v)\in \mathcal{E}_{LU}} w_{uv} d(h_\theta(x_u), h_\theta(x_v)) \nonumber \\ 
& \quad + \alpha_3 \sum_{(u,v)\in \mathcal{E}_{UU}} w_{uv} d(h_\theta(x_u), h_\theta(x_v), \label{eqn:cost_ngm}
\end{align}
where $\mathcal{E}_{LL}$, $\mathcal{E}_{LU}$, and $\mathcal{E}_{UU}$ are sets of labeled-labeled, labeled-unlabeled and unlabeled-unlabeled edges correspondingly, $h(\cdot)$ represents the hidden representations of the inputs produced by the neural network, and $d(\cdot)$ is a distance metric, and $\{\alpha_1, \alpha_2, \alpha_3\}$ are hyperparameters. Note that we have separated the terms based on the edge types, as these can affect the training differently. 

In practice, we choose an $l$-1 or $l$-2 distance metric for $d(\cdot)$, and $h(x)$ to be the last layer of the neural network. However, these choices can be changed, to a customized metric, or to using an intermediate hidden layer instead.

\subsection{Connections to previous methods}
The graph-dependent $\alpha$ hyperparameters control the balance of the contributions of different edge types. When $\{\alpha_i=0\}_{i=1}^{3}$, the proposed objective ignores the similarity constraint and becomes a supervised-only neural network objective as in equation \ref{eqn:cost_nn}. When only $\alpha_1 \neq 0$, the training cost has an additional term for labeled nodes, that acts as a regularizer. When $g_\theta(x) = h_\theta(x) = \hat{y}$, where $\hat{y}$ is the label distribution, the individual cost functions ($c$ and $d$) are squared $l$-2 norm, and the objective is trained using $\hat{y}$ directly instead of $\theta$, we arrive at the label propagation objective in equation \ref{eqn:cost_lp}. Therefore, the proposed objective could be thought of as a {\it non-linear} version of the label propagation objective, and a {\it graph-regularized} version of the neural network training objective. 

\subsection{Network inputs and graph construction\label{sec:graph_construction}}
Similar to graph-based label propagation, the choice of the input graphs is critical, to correctly bias the neural network's prediction. Depending on the type of the graphs and nodes in the graph, they can be readily available to use such as social networks or protein linking networks, or they can be constructed (a)~using generic graphs such as Knowledge Bases, that consists of relationship links between entities, (b)~using embeddings learnt by an unsupervised learning technique, or, (c)~using sparse feature representations for each vertex. Additionally, the proposed training objective can be easily modified for directed graphs.

\begin{figure}[!ht]
\vspace{-10pt}
\centering
\includegraphics[width=0.4\textwidth]{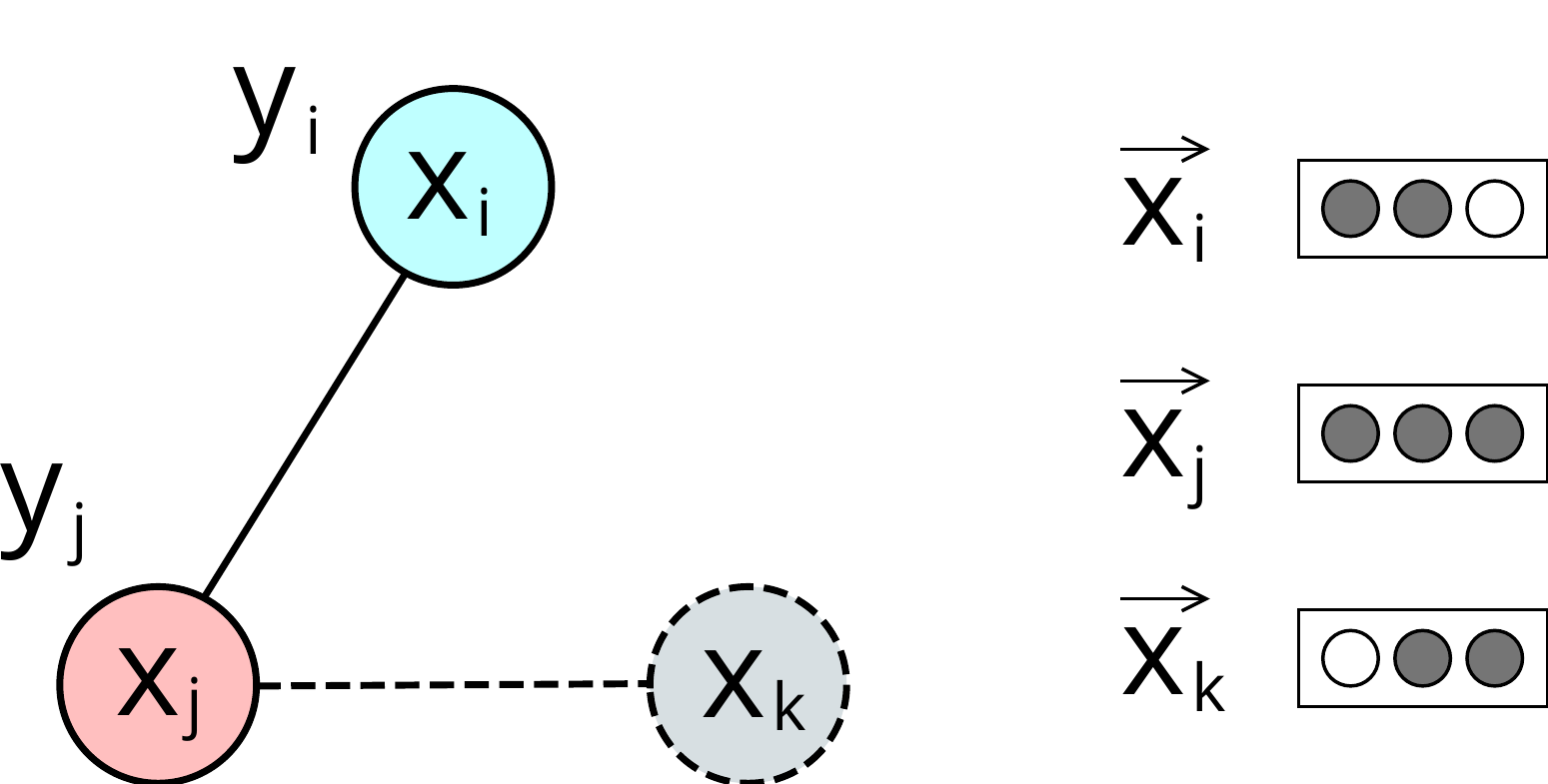}
\caption{Illustration of how we can construct inputs to the neural network using the adjacency matrix. In this example, we have three nodes and two edges. The feature vector created for each node {\it (shown on the right)} has 1's at its index and indices of nodes that it's adjacent to.\label{fig:adj}}
\end{figure}

We have discussed using node-level features as inputs to the neural network. In the absence of such inputs, our training scheme can still be deployed using input features derived from the graph itself. We show in figure \ref{fig:adj} and in experiments that the neighbourhood information such as rows in the adjacency matrix are simple to construct, yet powerful inputs to the network. These features can also be combined with existing features. 

When the number of graph nodes is high, this construction can have a high complexity and result in a large number of input features. This can be avoided by several ways: (i)~clustering the nodes and using the cluster assignments and similarities, (ii)~learning an embedding function of nodes \cite{perozzi2014deepwalk}, or (iii)~sampling the neighbourhood/context \cite{yang2016revisiting}. In practice, we observe that the input space can be bounded by a constant, even for massive graphs, with efficient scalable methods like unsupervised propagation (i.e., propagating node identity labels across the graph and selecting ones with highest support as input features to neural graph machines).

\subsection{Optimization}
The proposed objective function in equation \ref{eqn:cost_ngm} has several summations over the labeled points and edges, and can be equivalently written as follows,
\begin{align}
\mathcal{C}_{\mathrm{NGM}}(\theta) &= \sum_{(u,v)\in \mathcal{E}_{LL}} \alpha_1 w_{uv} d(h_\theta(x_u), h_\theta(x_v)) + c_{uv} \nonumber \\ & \;\;\; + \sum_{(u,v)\in \mathcal{E}_{LU}} \alpha_2 w_{uv} d(h_\theta(x_u), h_\theta(x_v)) + c_{u} \nonumber \\ & \;\;\;+ \sum_{(u,v)\in \mathcal{E}_{UU}} \alpha_3 w_{uv} d(h_\theta(x_u), h_\theta(x_v) \label{eqn:cost_ngm_new},
\end{align}
where
\begin{align}
c_{uv} &= \frac{1}{|u|} c (g_\theta(x_u), y_u) + \frac{1}{|v|}c (g_\theta(x_v), y_v) \nonumber\\
c_{u} &= \frac{1}{|u|} c (g_\theta(x_u), y_u), \nonumber
\end{align}
$|u|$ and $|v|$ are the number of edges incident to vertices $u$ and $v$, respectively. The objective in its new form enables stochastic training to be deployed by sampling edges. In particular, in each training iteration, we use a minibatch of edges and obtain the stochastic gradients of the objective. To further reduce noise and speedup learning, we sample edges within a neighbourhood region, that is to make sure some sampled edges have shared end nodes.

\subsection{Complexity}
The complexity of each epoch in training using equation \ref{eqn:cost_ngm_new} is $\mathcal{O}(M)$ where $M=|\mathcal{E}|$ is the number of edges in the graph. In the case where there is a large number of unlabeled-unlabeled edges, they potentially do not help learning and could be ignored, leading to a lower complexity. One strategy to include them is self-training, that is to grow seeds or labeled nodes as we train the networks. We experimentally demonstrate this technique in \cref{sec:doc_class}. Predictions at inference time can be made at the same cost as that of vanilla neural networks.

\section{Experiments}
\label{sec:experiments}
In this section, we provide several experiments showing the efficacy of the proposed training objective on a wide range of tasks, datasets and network architectures. All the experiments are done using a TensorFlow implementation \citep{tensorflow2015-whitepaper}.

\subsection{Multi-label Classification of Nodes on Graphs}
\label{sec:exp_graph}
We first demonstrate our approach using a multi-label classification problem on nodes in a relationship graph. 
In particular, the {\it BlogCatalog} dataset \citep{agarwal2009social}, a network of social relationships between bloggers is considered. 
This graph has 10,312 nodes, 333,983 edges and 39 labels per node, which represent the bloggers, their social connections and the bloggers' interests, respectively. 
Following previous approaches in the literature \citep{grover2016node2vec, agarwal2009social}, we train and make predictions using multiple one-vs-rest classifiers.

Since there are no provided features for each node, we use the rows of the adjacency matrix as input features, as discussed in section \ref{sec:graph_construction}. 
Feed-forward neural networks (FFNNs) with one hidden layer of 50 units are employed to map the constructed inputs to the node labels. 
As we use the test set to construct the graph and augment the training objective, the training in this experiment is transductive. 
Critically, to combat the unbalanced training set, we employ weighted sampling during training, i.e.~making sure each minibatch has both positive and negative examples. 
In this experiment, we fix $\alpha_i$ to be equal, and experiment with $\alpha=0.1$ and use the $l$-2 metric to compute the distance $d$ between the hidden representations of the networks. 
In addition, we create a range of train/test splits by varying the number of training points being presented to the networks.

We compare our method (NGM-FFNN) against a two-stage approach that first uses node2vec \citep{grover2016node2vec} to generate node embeddings and then uses a linear one-vs-rest classifier for classification. 
The methods are evaluated using two metrics Macro F1 and Micro F1. 
The average results for different train/test splits using our method and the baseline are included in table \ref{tab:mlp}. In addition, we compare NGM-FFNN with a non-augmented FFNN in which $\alpha=0$, i.e.~no edge information is used during training.
We observe that the graph-augmented training scheme performs better (6\% relative improvement on Macro F1 when the training set size is 20\% and 50\% of the dataset) or comparatively (when the training size is 80\%) compared to the vanilla neural networks trained with no edge information.
Both methods significantly outperform the approach that uses node embeddings and linear classifiers.
We observe the same improvement over node2vec on the Micro F1 metric and NGM-FFNN is comparable to vanilla FFNN ($\alpha=0$) but outperforms other methods on the recall metric.
\begin{table}[!ht]
\centering
\caption{Macro F1 results for BlogCatalog dataset averaged over 10 random splits. The higher is better. Graph regularized neural networks outperform  node2vec embedding and a linear classifer in all training size settings.\label{tab:mlp}}
\vspace{5pt}
\begin{tabular}{|c|c c|}
\toprule
$|$Train$|$ / $|$Dataset$|$ & NGM-FFNN & node2vec\footnotemark{}\\
\midrule
20\% & \bf 0.191 & 0.168 \\
50\% & \bf 0.242 & 0.174 \\
80\% & \bf 0.262 & 0.177 \\
\bottomrule
\end{tabular}
\end{table}
\footnotetext{These results are different compared to \cite{grover2016node2vec}, since we treat the classifiers (one per label) independently. Both methods shown here use the exact same setting and training/test data splits.}

% & $\alpha = 0$ 
% & 0.180 &
 % & 0.238 &
 % & \bf 0.263

These results demonstrate that using the graph itself as direct inputs to the neural network and letting the network figure out a non-linear mapping directly from the raw graph is more effective than the two-stage approach considered. More importantly, the results also show that using the graph information improves the performance in the limited data regime (for example: when training set is only 20\% or 50\% of the dataset). 

\subsection{Text Classification using Character-level CNNs}
\label{sec:exp_cnn}
We evaluate the proposed objective function on a multi-class text classification task using a character-level convolutional neural network (CNN). We use the AG news dataset from \cite{zhang2015text}, where the task is to classify a news article into one of 4 categories. Each category has 30,000 examples for training and 1,900 examples for testing. In addition to the train and test sets, there are 111,469 examples that are treated as unlabeled examples. 

As there is no provided graph structure linking the articles, we create such a graph based on the embeddings of the articles. We restrict the graph construction to only the train set and the unlabeled examples and keep the test set only for evaluation. We use the Google News word2vec corpus to calculate the average embedding for each news article and use the cosine similarity of document embeddings as a similarity metric. Each node is restricted to have a maximum of 5 neighbors.

We construct the CNN in the same way as \citep{zhang2015text} and pick their competitive {\it ``small CNN''} as our baseline for a more reasonable comparison to our set-up. Our approach employs the same network, but with significantly smaller number of convolutional layers and layer sizes, as shown in table \ref{tab:cnn_config}.
\begin{table}[!ht]
\centering
\caption{Settings of CNNs for the text classification experiment, including the number of convolutional layers and their sizes. The baseline model is the {\it small CNN} from \cite{zhang2015text} and is significantly larger than our model.\label{tab:cnn_config}}
\vspace{5pt}
\begin{tabular}{|c|c|c|}
\toprule
Setting & Baseline &Our ``tiny CNN'' \\
\midrule
\# of conv. layers & 6 & 3 \\
Frame size in conv.~layers & 256 & 32 \\
\# of FC layers & 3 & 3 \\
Hidden units in FC layers & 1024 & 256 \\
\bottomrule
\end{tabular}
\end{table}

The networks are trained with the same hyper-parameters as reported in \cite{zhang2015text}. We observed that the model converged within 20 epochs (the model loss did not change much) and hence used this as a stopping criterion for this task. Experiments also showed that running the network for longer also did not change the qualitative performance. We use the cross entropy loss on the final outputs of the network, that is $d = \mathrm{cross\_entropy}(g(x_u), g(x_v))$, to compute the distance between nodes on an edge. In addition, we also experiment with a data augmentation technique using an English thesaurus, as done in \cite{zhang2015text}.

We compare the ``tiny CNN'' trained using the proposed objective function with the baseline using the accuracy on the test set in table \ref{tab:cnn}. Our approach outperforms the baseline by provides a 1.8\% absolute and 2.1\% relative improvement in accuracy, despite using a much smaller network. In addition, our model with graph augmentation trains much faster and produces results on par or better than the performance of a significantly larger network, {\it ``large CNN"} \cite{zhang2015text}, which has an accuracy of 87.18 without using a thesaurus, and 86.61 with the thesaurus.

\begin{table}[!ht]
\centering
\caption{Results for news article categorization using character-level CNNs. Our method gives better predictive accuracy, despite using a much smaller CNN compared to the ``small CNN'' baseline from \cite{zhang2015text}$^\ddagger$.\label{tab:cnn}}
\vspace{5pt}
\begin{tabular}{|c|c|}
\toprule
Network & Accuracy \% \\
\midrule
Baseline$^\ddagger$ & 84.35 \\
Baseline with thesaurus augmentation$^\ddagger$ & 85.20 \\ \hline
Our ``tiny'' CNN & 85.07 \\ 
Our ``tiny'' CNN with NGM& \bf 86.90 \\
\bottomrule
\end{tabular}
\end{table}

\subsection{Semantic Intent Classification using LSTM RNNs}
\label{sec:exp_rnn}
We compare the performance of our approach for training RNN sequence models (LSTM) for a semantic intent classification task as described in the recent work on SmartReply \citep{smartreply2016} for automatically generating short email responses. One of the underlying tasks in SmartReply is to discover and map short response messages to semantic intent clusters.\footnote{For details regarding SmartReply and how the semantic intent clusters are generated, refer \cite{smartreply2016}.} We choose 20 intent classes and created a dataset comprised of 5,483 samples (3,832 for training, 560 for validation and 1,091 for testing). Each sample instance corresponds to a short response message text paired with a semantic intent category that was manually verified by human annotators. For example, {\it``That sounds awesome!''} and {\it``Sounds fabulous''} belong to the {\it sounds good} intent cluster. 

We construct a sparse graph in a similar manner as the news categorization task using word2vec embeddings over the message text and computing similarity to generate a response message graph with fixed node degree (k=10). We use $l$-2 for the distance metric $d(\cdot)$ and choose $\alpha$ based on the development set.

We run the experiments for a fixed number of time steps and pick the best results on the development set. A multilayer LSTM architecture (2 layers, 100 dimensions) is used for the RNN sequence model. The LSTM model and its NGM variant are also compared against other baseline systems---{\it Random} baseline ranks the intent categories randomly and {\it Frequency} baseline ranks them in order of their frequency in the training corpus. To evaluate the intent prediction quality of different approaches, for each test instance, we  compute the rank of the actual intent category $\mathrm{rank}_i$ with respect to the ranking produced by the method and use this to calculate the Mean Reciprocal Rank:\\
\begin{equation*}
\nonumber \mathrm{MRR}=\frac{1}{N} \sum_{i=1}^N \frac{1}{\mathrm{rank}_i} 
\end{equation*}

We show in table \ref{tab:rnn} that LSTM RNNs with our proposed graph-augmented training objective function outperform standard baselines by achieving a better MRR. 

\begin{table}[!ht]
\centering
\caption{Results for Semantic Intent Classification using graph-augmented LSTM RNNs and baselines. Higher MRR is better.\label{tab:rnn}}
\vspace{5pt}
\begin{tabular}{|c|c|}
\toprule
Model & Mean Reciprocal Rank (MRR) \\
\midrule
Random & 0.175  \\ \hline
Frequency & 0.258  \\ \hline
LSTM & 0.276 \\ \hline
NGM-LSTM & \bf 0.284 \\
\bottomrule
\end{tabular}
\end{table}

\subsection{Low-supervision Document Classification}
\label{sec:doc_class}
Finally, we compare our method on a task with very limited supervision---the PubMed document classification problem \cite{sen2008collective}. The task is to classify each document into one of 3 classes, with each document being described by a TF-IDF weighted word vector. The graph is available as a citation network: two documents are connected to each other if one cites the other. The graph has 19,717 nodes and 44,338 edges, with each class having 20 seed nodes and 1000 test nodes. In our experiments we exclude the test nodes from the graph entirely, training only on the labeled and unlabeled nodes.

We train a feed-forward neural network (FFNN) with two hidden layers with 250 and 100 neurons, using the $l$-2 distance metric on the last hidden layer. The NGM-FFNN model is trained with $\alpha_i = 0.2$, while the baseline FFNN is trained with $\alpha_i = 0$ (i.e., a supervised-only model). We use self-training to train the model, starting with just the 60 seed nodes (20 per class) as training data. The amount of training data is iteratively increased by assigning labels to the immediate neighbors of the labeled nodes and retraining the model. For the self-trained NGM-FFNN model, this strategy results in incrementally growing the neighborhood and thereby, $LL$ and $LU$ edges in equation \ref{eqn:cost_ngm_new} objective. 

We compare the final NGM-FFNN model against the FFNN baseline and other techniques reported in \cite{yang2016revisiting} including the Planetoid models \cite{yang2016revisiting}, semi-supervised embedding \cite{weston2012deep}, manifold regression \cite{belkin2006manifold}, transductive SVM \cite{joachims1999transductive}, label propagation \cite{zhu2003semi}, graph embeddings \cite{perozzi2014deepwalk} and a linear softmax model. Full results are included in \cref{tab:pubmed}. 
\vspace{-0.1in}

\begin{table}[!ht]
\centering
\caption{Results for document classification on the PubMed dataset using neural networks. The top results are taken from \cite{yang2016revisiting}. The bottom two rows are ours, with the NGM training outperforming all other baselines, except Planetoid-I. Please see text for relevant references.\label{tab:pubmed}}
\vspace{5pt}
\begin{tabular}{|c|c|}
\toprule
Method & Accuracy \\
\midrule
Linear + Softmax         & 0.698\\
Semi-supervised embedding      & 0.711
\\
Manifold regularization      & 0.707\\
Transductive SVM         & 0.622
\\
Label propagation           & 0.630\\

Graph embedding     & 0.653
\\

Planetoid-I  & 0.772\\
Planetoid-G  & 0.664
\\
Planetoid-T  & 0.757\\
\midrule
Feed-forward NN & 0.709
\\
NGM-FFNN           & 0.759\\
\bottomrule
\end{tabular}
\end{table}
\vspace{-0.1in}
The results show that the NGM model (without any tuning) outperforms many baselines including FFNN, semi-supervised embedding, manifold regularization and Planetoid-G/Planetoid-T, and compares favorably to Planetoid-I. Most importantly, this result demonstrates the graph augmentation scheme can lead to better regularised neural networks, especially in low sample regime (20 samples per class in this case). We believe that with tuning, NGM accuracy can be improved even further.

\section{Conclusions}
\label{sec:summary}

We have revisited graph-augmentation training of neural networks and proposed Neural Graph Machines as a general framework for doing so. Its objective function encourages the neural networks to make accurate node-level predictions, as in vanilla neural network training, as well as constrains the networks to learn similar hidden representations for nodes connected by an edge in the graph. Importantly, the objective can be trained by stochastic gradient descent and scaled to large graphs.

We validated the efficacy of the graph-augmented objective on various tasks including bloggers' interest, text category and semantic intent classification problems, using a wide range of neural network architectures (FFNNs, CNNs and LSTM RNNs). The experimental results demonstrated that graph-augmented training almost always helps to find better neural networks that outperforms other techniques in predictive performance or even much smaller networks that are faster and easier to train. Additionally, the node-level input features can be combined with graph features as inputs to the neural networks. We showed that a neural network that simply takes the adjacency matrix of a graph and produces node labels, can perform better than a recently proposed two-stage approach using sophisticated graph embeddings and a linear classifier. Our framework also excels when the neural network is small, or when there is limited supervision available.

While our objective can be applied to multiple graphs which come from different domains, we have not fully explored this aspect and leave this as future work. We expect the domain-specific networks can interact with the graphs to determine the importance of each domain/graph source in prediction. We also did not explore using graph regularisation for different hidden layers of the neural networks; we expect this is key for the multi-graph transfer setting \cite{yosinski2014transferable}. Another possible future extension is to use our objective on directed graphs, that is to control the direction of influence between nodes during training.

\section*{Acknowledgements} 
We would like to thank the Google Expander team for insightful feedback.

\bibliography{neural-graph}

\begin{thebibliography}{22}
\providecommand{\natexlab}[1]{#1}
\providecommand{\url}[1]{\texttt{#1}}
\expandafter\ifx\csname urlstyle\endcsname\relax
  \providecommand{\doi}[1]{doi: #1}\else
  \providecommand{\doi}{doi: \begingroup \urlstyle{rm}\Url}\fi

\bibitem[Abadi et~al.(2015)Abadi, Agarwal, Barham, Brevdo, Chen, Citro,
  Corrado, Davis, Dean, Devin, Ghemawat, Goodfellow, Harp, Irving, Isard, Jia,
  Jozefowicz, Kaiser, Kudlur, Levenberg, Man\'{e}, Monga, Moore, Murray, Olah,
  Schuster, Shlens, Steiner, Sutskever, Talwar, Tucker, Vanhoucke, Vasudevan,
  Vi\'{e}gas, Vinyals, Warden, Wattenberg, Wicke, Yu, and
  Zheng]{tensorflow2015-whitepaper}
Abadi, Mart\'{\i}n, Agarwal, Ashish, Barham, Paul, Brevdo, Eugene, Chen,
  Zhifeng, Citro, Craig, Corrado, Greg~S., Davis, Andy, Dean, Jeffrey, Devin,
  Matthieu, Ghemawat, Sanjay, Goodfellow, Ian, Harp, Andrew, Irving, Geoffrey,
  Isard, Michael, Jia, Yangqing, Jozefowicz, Rafal, Kaiser, Lukasz, Kudlur,
  Manjunath, Levenberg, Josh, Man\'{e}, Dan, Monga, Rajat, Moore, Sherry,
  Murray, Derek, Olah, Chris, Schuster, Mike, Shlens, Jonathon, Steiner,
  Benoit, Sutskever, Ilya, Talwar, Kunal, Tucker, Paul, Vanhoucke, Vincent,
  Vasudevan, Vijay, Vi\'{e}gas, Fernanda, Vinyals, Oriol, Warden, Pete,
  Wattenberg, Martin, Wicke, Martin, Yu, Yuan, and Zheng, Xiaoqiang.
\newblock {TensorFlow}: Large-scale machine learning on heterogeneous systems,
  2015.
\newblock URL \url{http://tensorflow.org/}.
\newblock Software available from tensorflow.org.

\bibitem[Agarwal et~al.(2009)Agarwal, Liu, Murthy, Sen, and
  Wang]{agarwal2009social}
Agarwal, Nitin, Liu, Huan, Murthy, Sudheendra, Sen, Arunabha, and Wang, Xufei.
\newblock A social identity approach to identify familiar strangers in a social
  network.
\newblock In \emph{3rd International AAAI Conference on Weblogs and Social
  Media (ICWSM09)}, 2009.

\bibitem[Belkin et~al.(2006)Belkin, Niyogi, and Sindhwani]{belkin2006manifold}
Belkin, Mikhail, Niyogi, Partha, and Sindhwani, Vikas.
\newblock Manifold regularization: A geometric framework for learning from
  labeled and unlabeled examples.
\newblock \emph{Journal of machine learning research}, 7\penalty0
  (Nov):\penalty0 2399--2434, 2006.

\bibitem[Bengio et~al.(2006)Bengio, Delalleau, and Le~Roux]{bengio2006label}
Bengio, Yoshua, Delalleau, Olivier, and Le~Roux, Nicolas.
\newblock Label propagation and quadratic criterion.
\newblock In Chapelle, O, Scholkopf, B, and Zien, A (eds.),
  \emph{Semi-supervised learning}, pp.\  193--216. MIT Press, 2006.

\bibitem[Defferrard et~al.(2016)Defferrard, Bresson, and
  Vandergheynst]{defferrard2016convolutional}
Defferrard, Micha{\"e}l, Bresson, Xavier, and Vandergheynst, Pierre.
\newblock Convolutional neural networks on graphs with fast localized spectral
  filtering.
\newblock In \emph{Advances in Neural Information Processing Systems}, pp.\
  3837--3845, 2016.

\bibitem[Grover \& Leskovec(2016)Grover and Leskovec]{grover2016node2vec}
Grover, Aditya and Leskovec, Jure.
\newblock node2vec: Scalable feature learning for networks.
\newblock In \emph{Proceedings of the 22nd {ACM} {SIGKDD} International
  Conference on Knowledge Discovery and Data Mining, San Francisco, CA, USA,
  August 13-17, 2016}, pp.\  855--864, 2016.

\bibitem[Hinton et~al.(2012)Hinton, Deng, Yu, Dahl, Mohamed, Jaitly, Senior,
  Vanhoucke, Nguyen, Sainath, et~al.]{hinton2012deep}
Hinton, Geoffrey, Deng, Li, Yu, Dong, Dahl, George~E, Mohamed, Abdel-rahman,
  Jaitly, Navdeep, Senior, Andrew, Vanhoucke, Vincent, Nguyen, Patrick,
  Sainath, Tara~N, et~al.
\newblock Deep neural networks for acoustic modeling in speech recognition: The
  shared views of four research groups.
\newblock \emph{IEEE Signal Processing Magazine}, 29\penalty0 (6):\penalty0
  82--97, 2012.

\bibitem[Joachims(1999)]{joachims1999transductive}
Joachims, Thorsten.
\newblock Transductive inference for text classification using support vector
  machines.
\newblock In \emph{International Conference on Machine Learning}, 1999.

\bibitem[Kannan et~al.(2016)Kannan, Kurach, Ravi, Kaufmann, Tomkins, Miklos,
  Corrado, Lukacs, Ganea, Young, and Ramavajjala]{smartreply2016}
Kannan, Anjuli, Kurach, Karol, Ravi, Sujith, Kaufmann, Tobias, Tomkins, Andrew,
  Miklos, Balint, Corrado, Greg, Lukacs, Laszlo, Ganea, Marina, Young, Peter,
  and Ramavajjala, Vivek.
\newblock Smart reply: Automated response suggestion for email.
\newblock In \emph{Proceedings of the ACM SIGKDD Conference on Knowledge
  Discovery and Data Mining (KDD)}, 2016.

\bibitem[Kingma et~al.(2014)Kingma, Mohamed, Rezende, and
  Welling]{kingma2014semi}
Kingma, Diederik~P, Mohamed, Shakir, Rezende, Danilo~Jimenez, and Welling, Max.
\newblock Semi-supervised learning with deep generative models.
\newblock In \emph{Advances in Neural Information Processing Systems}, pp.\
  3581--3589, 2014.

\bibitem[Krizhevsky et~al.(2012)Krizhevsky, Sutskever, and
  Hinton]{krizhevsky2012imagenet}
Krizhevsky, Alex, Sutskever, Ilya, and Hinton, Geoffrey~E.
\newblock Imagenet classification with deep convolutional neural networks.
\newblock In \emph{Advances in Neural Information Processing Systems}, pp.\
  1097--1105, 2012.

\bibitem[Lee(2013)]{lee2013pseudo}
Lee, Dong-Hyun.
\newblock Pseudo-label: The simple and efficient semi-supervised learning
  method for deep neural networks.
\newblock In \emph{ICML 2013 Workshop : Challenges in Representation Learning
  (WREPL)}, 2013.

\bibitem[Perozzi et~al.(2014)Perozzi, Al-Rfou, and Skiena]{perozzi2014deepwalk}
Perozzi, Bryan, Al-Rfou, Rami, and Skiena, Steven.
\newblock Deepwalk: Online learning of social representations.
\newblock In \emph{Proceedings of the 20th ACM SIGKDD international conference
  on Knowledge discovery and data mining}, pp.\  701--710. ACM, 2014.

\bibitem[Ravi \& Diao(2016)Ravi and Diao]{ravi2016large}
Ravi, Sujith and Diao, Qiming.
\newblock Large scale distributed semi-supervised learning using streaming
  approximation.
\newblock In \emph{Proceedings of the 19th International Conference on
  Artificial Intelligence and Statistics}, pp.\  519--528, 2016.

\bibitem[Sen et~al.(2008)Sen, Namata, Bilgic, Getoor, Galligher, and
  Eliassi-Rad]{sen2008collective}
Sen, Prithviraj, Namata, Galileo, Bilgic, Mustafa, Getoor, Lise, Galligher,
  Brian, and Eliassi-Rad, Tina.
\newblock Collective classification in network data.
\newblock \emph{AI magazine}, 29\penalty0 (3):\penalty0 93, 2008.

\bibitem[Sutskever et~al.(2014)Sutskever, Vinyals, and
  Le]{sutskever2014sequence}
Sutskever, Ilya, Vinyals, Oriol, and Le, Quoc~V.
\newblock Sequence to sequence learning with neural networks.
\newblock In \emph{Advances in Neural Information Processing Systems}, pp.\
  3104--3112, 2014.

\bibitem[Weston et~al.(2012)Weston, Ratle, Mobahi, and
  Collobert]{weston2012deep}
Weston, Jason, Ratle, Fr{\'e}d{\'e}ric, Mobahi, Hossein, and Collobert, Ronan.
\newblock Deep learning via semi-supervised embedding.
\newblock In \emph{Neural Networks: Tricks of the Trade}, pp.\  639--655.
  Springer, 2012.

\bibitem[Yang et~al.(2016)Yang, Cohen, and Salakhudinov]{yang2016revisiting}
Yang, Zhilin, Cohen, William, and Salakhudinov, Ruslan.
\newblock Revisiting semi-supervised learning with graph embeddings.
\newblock In \emph{Proceedings of The 33rd International Conference on Machine
  Learning}, pp.\  40--48, 2016.

\bibitem[Yosinski et~al.(2014)Yosinski, Clune, Bengio, and
  Lipson]{yosinski2014transferable}
Yosinski, Jason, Clune, Jeff, Bengio, Yoshua, and Lipson, Hod.
\newblock How transferable are features in deep neural networks?
\newblock In \emph{Advances in Neural Information Processing Systems}, pp.\
  3320--3328, 2014.

\bibitem[Zhang et~al.(2015)Zhang, Zhao, and LeCun]{zhang2015text}
Zhang, Xiang, Zhao, Junbo, and LeCun, Yann.
\newblock Character-level convolutional networks for text classification.
\newblock In \emph{Advances in Neural Information Processing Systems}, pp.\
  649--657, 2015.

\bibitem[Zhu et~al.(2003)Zhu, Ghahramani, and Lafferty]{zhu2003semi}
Zhu, X, Ghahramani, Z, and Lafferty, J.
\newblock Semi-supervised learning using gaussian fields and harmonic
  functions.
\newblock In \emph{Proceedings of the 20th International Conference on Machine
  Learning (ICML-2003) Volume 2}, volume~2, pp.\  912--919. AIAA Press, 2003.

\bibitem[Zhu \& Ghahramani()Zhu and Ghahramani]{zhu2002learning}
Zhu, Xiaojin and Ghahramani, Zoubin.
\newblock Learning from labeled and unlabeled data with label propagation.
\newblock Technical report, School of Computer Science, Canegie Mellon
  University.

\end{thebibliography}
\bibliographystyle{icml2017}

\end{document}